\documentclass{article}

    \usepackage[preprint]{neurips_2025}

\usepackage{comment}
\usepackage[utf8]{inputenc} % allow utf-8 input
\usepackage[T1]{fontenc}    % use 8-bit T1 fonts
\usepackage{hyperref}       % hyperlinks
\usepackage{url}            % simple URL typesetting
\usepackage{booktabs}       % professional-quality tables
\usepackage{amsfonts}       % blackboard math symbols
\usepackage{nicefrac}       % compact symbols for 1/2, etc.
\usepackage{microtype}      % microtypography
\usepackage{xcolor}         % colors
\usepackage{wrapfig}
\usepackage{graphicx}
\usepackage{subcaption}
\def\code#1{\texttt{#1}}

\usepackage{titlesec}
\titlespacing*{\paragraph}{0pt}{1ex}{1em}

\title{Multi-Agent Craftax: Benchmarking Open-Ended Multi-Agent Reinforcement Learning \\ at the Hyperscale}

\author{%
  Bassel Al Omari\textsuperscript{1}\thanks{Work done while visiting FLAIR. Correspondence to \texttt{b2alomar@uwaterloo.ca}.} \quad
  Michael Matthews\textsuperscript{2} \quad
  Alexander Rutherford\textsuperscript{2} \quad
  Jakob N. Foerster\textsuperscript{2} \\
  \textsuperscript{1}University of Waterloo \\
  \textsuperscript{2}FLAIR, University of Oxford
}

\begin{document}

\maketitle

\begin{abstract}
Progress in multi-agent reinforcement learning (MARL) requires challenging benchmarks that assess the limits of current methods. However, existing benchmarks often target narrow short-horizon challenges that do not adequately stress the long-term dependencies and generalization capabilities inherent in many multi-agent systems. To address this, we first present \textit{Craftax-MA}: an extension of the popular open-ended RL environment, Craftax, that supports multiple agents and evaluates a wide range of general abilities within a single environment. Written in JAX, \textit{Craftax-MA} is exceptionally fast with a training run using 250 million environment interactions completing in under an hour. To provide a more compelling challenge for MARL, we also present \textit{Craftax-Coop}, an extension introducing heterogeneous agents, trading and more mechanics that require complex cooperation among agents for success\setcounter{footnote}{0}\footnote{Code is available at \url{https://github.com/BaselOmari/MA-Craftax}.}. We provide analysis demonstrating that existing algorithms struggle with key challenges in this benchmark, including long-horizon credit assignment, exploration and cooperation, and argue for its potential to drive long-term research in MARL.
\end{abstract}

\section{Introduction}
\begin{figure}[t]
    \centering
    \begin{subfigure}[c]{0.45\textwidth}
        \includegraphics[width=\linewidth]{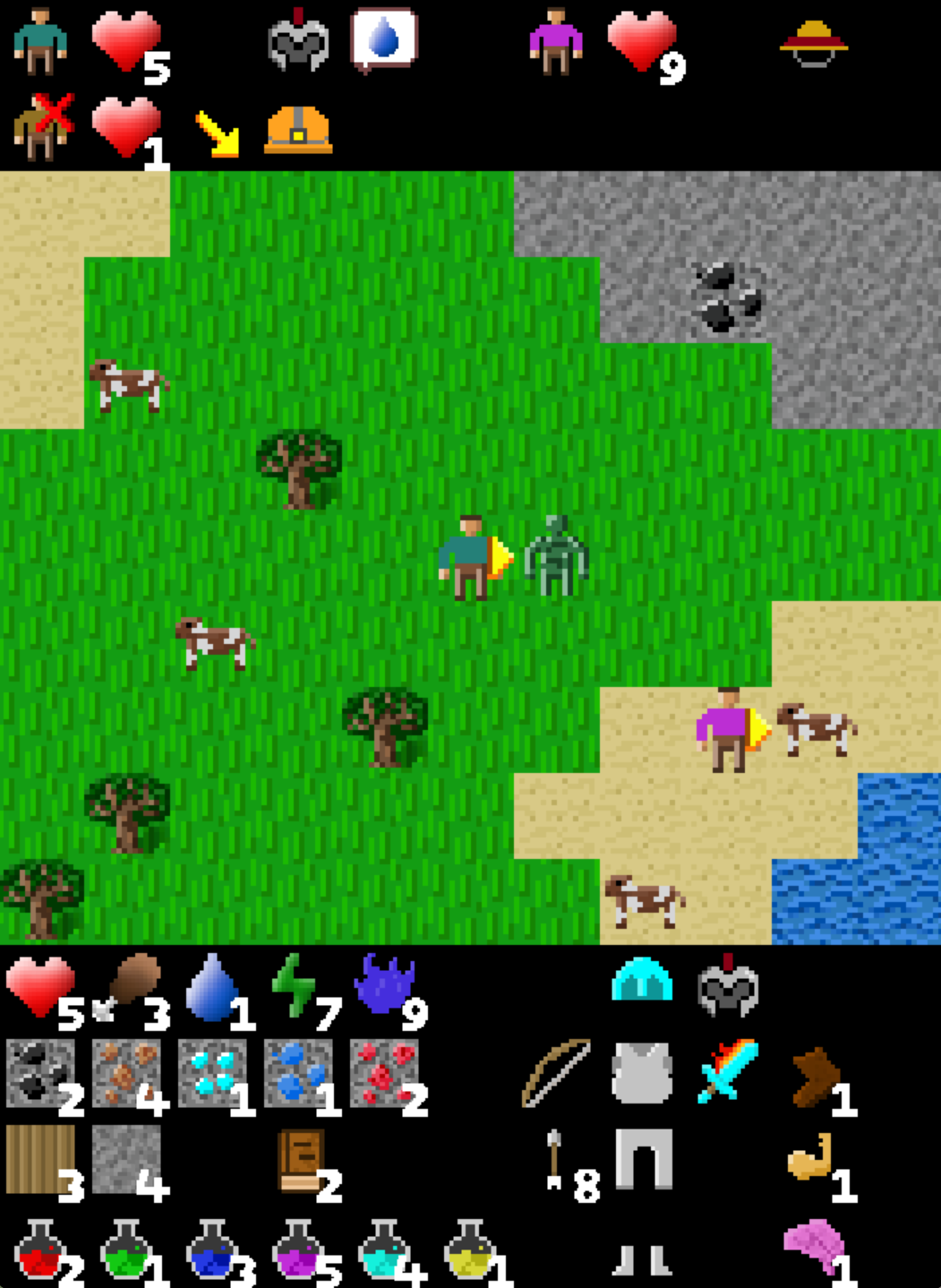}
        % \caption{Pixel Observation}
        \label{fig:pixel_obs}
    \end{subfigure}
    \hspace{0.02\textwidth}
    \begin{subfigure}[c]{0.49\textwidth}
        \includegraphics[width=\linewidth]{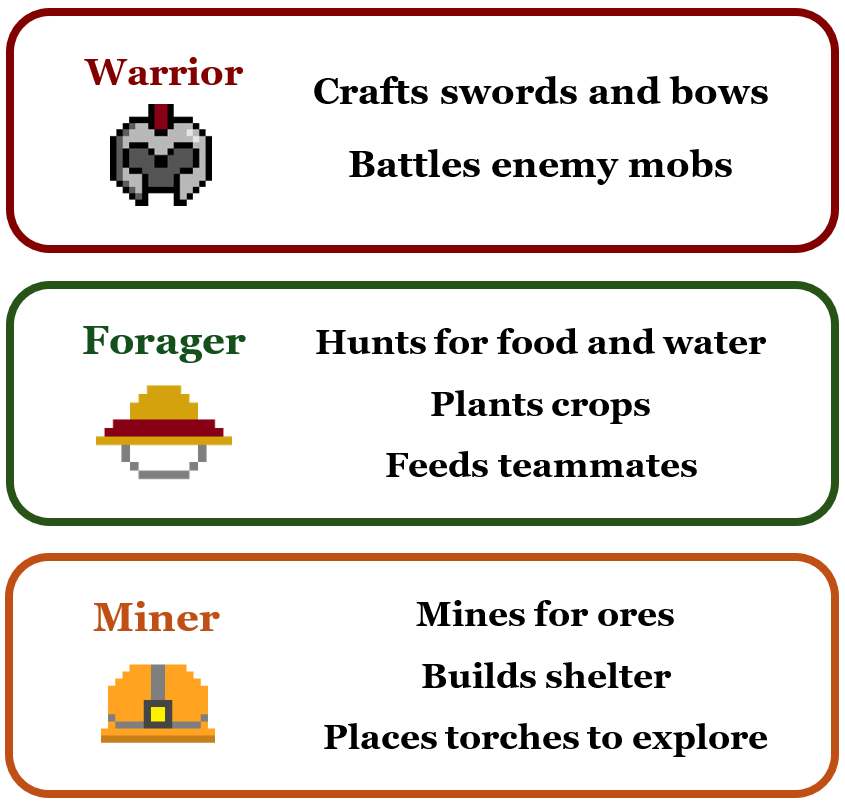}
        % \caption{Specializations}
        \label{fig:specializations}
    \end{subfigure}
    \caption{Example pixel-based observation of \textit{Craftax-Coop} with a summary of player specializations. We also provide a symbolic observation to focus research on multi-agent challenges.}
    \label{fig:craftax_summary}
\end{figure}
Progress in reinforcement learning (RL) goes hand in hand with the development of increasingly sophisticated environments. Such environments serve as benchmarks focusing research towards well-defined problems and enabling direct comparison between methods. In the effort towards increasingly general agents, a community has arisen focused on environments with open-ended dynamics \citep{stanley2017openendedness}. These dynamics, such as procedurally generated worlds as seen in environments like Procgen \citep{cobbe2020procgen} and NetHack \citep{kuettler2020nethack}, help evaluate the systematic generalization of RL methods across diverse scenarios. This field has further advanced with the introduction of hardware-accelerated benchmarks, such as Craftax \citep{matthews2024craftax}, enabling rapid evaluation of scalable and adaptable RL methods using minimal computational resources. 

Despite these advancements in the single agent setting, the transition of these explorations to the multi-agent setting has remained limited. Existing multi-agent reinforcement learning (MARL) benchmarks often focus on narrow challenges under short time-horizons, such as micromanagement tasks in the popular SMAC benchmark \citep{samvelyan2019smac} or cooperation under imperfect information in Hanabi \citep{bard2020hanabi}. While such benchmarks have driven significant algorithmic development, their limited scopes prevent them from capturing the rich dependencies inherent in open-ended multi-agent systems with extended time horizons.

To address this gap, we present \textit{Craftax-MA}, a multi-agent extension of the Craftax benchmark. This extension preserves the rich dynamics of the original Craftax, enabling agents to gather resources, craft advanced tools and combat enemies while learning to navigate complex procedurally generated worlds. By providing a flexible testbed, these dynamics support the study of diverse multi-agent interactions over long time horizons, including competition over resources, coordinated exploration and dynamic role allocation. Built with JAX, \textit{Craftax-MA} is also fast, enabling rapid experimentation of complex multi-agent behavior even with limited computational resources. 

Building on \textit{Craftax-MA}, we present \textit{Craftax-Coop}, an enhanced multi-agent environment designed specifically to test cooperation among agents. \textit{Craftax-Coop} introduces agent specialization, where each agent is assigned one of three unique roles (Forager, Miner, or Warrior) each with distinct abilities and responsibilities. Agents must trade essential resources, coordinate their actions and maintain their health through collaborative strategies. \textit{Craftax-Coop} is thus an ideal testbed for exploring complex cooperation, long horizon planning and emergent teamwork in MARL.

Through experimental results, we demonstrate that popular MARL adaptations of algorithms fail to solve \textit{Craftax-MA} and \textit{Craftax-Coop}, particularly struggling with long-horizon credit assignment, efficient exploration and cooperation among heterogeneous agents. In some of our settings naive independent learning beats the popular MARL adaptation of PPO, which shows that they are overfit to a small number of popular MARL environments. We believe our environment suite will serve as a much needed challenging and robust benchmark for future MARL research, driving the development of more adaptable and cooperative multi-agent systems.

In summary, our contributions are:
\begin{enumerate}
\item We introduce \textit{Craftax-MA}, a multi-agent extension of the popular open-ended RL environment Craftax, which allows for individual or shared reward settings.
\item We introduce \textit{Craftax-Coop}, a co-operative extension of \textit{Craftax-MA} featuring specialized agent roles, trading among other mechanics that require continual cooperation among agents.
\item We demonstrate that existing MARL algorithms achieve limited success in \textit{Craftax-MA} and \textit{Craftax-Coop}, struggling specifically with credit assignment, exploration and cooperation.
\end{enumerate}

\section{Background}
\subsection{Multi-Agent Reinforcement Learning}

Multi-Agent Reinforcement Learning (MARL) extends the RL paradigm to facilitate the co-learning of multiple agents simultaneously in a single environment. In particular, we focus on fully co-operative MARL, a paradigm formalized in the decentralized partially observable Markov decision process (Dec-POMDP) \citep{oliehoek2016concise}. This is defined as $\langle \mathcal{S}, \mathcal{A}, \mathcal{O}, \mathcal{R}, \mathcal{T}, n\rangle$, where $\mathcal{S}$ is the set of states; $\mathcal{A}$ is the set of actions shared between all agents; $\mathcal{O}$ is the observation function $\mathcal{O}(s, i)$, mapping a state and agent index to a local observation; $\mathcal{T}: \mathcal{S} \times \mathcal{A}^n \to \Delta \mathcal{S}$ is the transition function, defining the distribution over next states $\mathcal{T}(s, a_1, ..., a_n)$ given a current state $s$ and tuple of actions $(a_1, ..., a_n)$; $\mathcal{R}: \mathcal{S} \to \mathbb{R}$ is the shared reward function and $n$ is the number of agents. Note that, since the agents all receive the same reward, they are incentivized to behave entirely cooperatively. The Dec-POMDP can be further generalized to a partially observable stochastic game (POSG), where the reward function additionally conditions on the agent index, allowing for the representation of general-sum games.

We follow the centralized training decentralized execution (CTDE) paradigm \citep{oliehoek2008optimal, kraemer2016multi}, where it is assumed that data can be shared between agents at training time to facilitate learning (for example through shared critics \citep{foerster2018counterfactual, gupta2017cooperative}), but each agent must behave with only local information at execution time.

\subsection{Hardware-Accelerated Reinforcement Learning}

The recent advent of training pipelines entirely on a hardware accelerator \citep{hessel2021podracer, lu2022discovered} has driven a revolution in RL, allowing for experiments to be run on orders of magnitude more samples than was previously possible. The key to unlocking this scale is the development of simulation environments that can run natively on hardware accelerators \citep{freeman2021brax, gymnax2022github, bonnet2023jumanji, koyamada2023pgx, matthews2024craftax, matthews2024kinetix, pignatelli2024navix, kazemkhani2024gpudrive}, allowing for massive parallelization of workers and the elimination of CPU-GPU data transfer. The vast majority of these environments have been implemented in JAX \citep{jax2018github}.

\subsection{Craftax and Crafter}

Craftax \citep{matthews2024craftax} is a challenging, single-agent, JAX-based environment that takes inspiration from the original Crafter benchmark \citep{hafner2021benchmarking} and the NetHack Learning Environment \citep{kuttler2020nethack}. Beating the game requires navigating nine floors of increasingly difficult enemies, making of use of game mechanics like crafting, building and combat. In particular, Craftax tests the capability of algorithms to explore, generalize to new settings and to perform long-term reasoning over many thousands of timesteps. At the time of writing, the benchmark is currently unsolved, with the best performing agent \citep{transformerXL} averaging only 18\% of the maximum reward.

\section{Multi-Agent Craftax}

\subsection{Craftax-MA}
We first present \textit{Craftax-MA}, a rewrite of the Craftax environment that supports multiple agents. Craftax-MA retains all of the original dynamics in Craftax, providing a straightforward transition for those already familiar with Craftax or Crafter. As with the original environments, agents here must explore procedurally generated worlds, gather resources, craft advanced tools and combat enemies, evaluating their abilities in generalization, deep exploration and long-term reasoning. The environment can support an unbounded number of agents, providing a scalable benchmark for studying micro and macro-scale multi-agent interactions. Some changes to the original environment were made to accommodate multiple agents (see Appendix A for more details).

\subsection{Craftax-Coop}
Through human demonstration, \citet{matthews2024craftax} highlight that Craftax can be completed with a single player, reducing the need for multiple agents. To provide a more compelling challenge for MARL we present \textit{Craftax-Coop}, an extension of \textit{Craftax-MA} introducing new mechanics designed to require continual cooperation between agents for success. In this section, we provide a brief overview of these changes.

\paragraph{Trading} 
\textit{Craftax-Coop} features a flexible trading system, allowing agents to exchange acquired resources between their inventories. Agents broadcast a request for a needed resource, which can then be fulfilled by others with access to that resource. To facilitate more trading, agents can trade items at any distance between each other. Agents are only permitted to trade base materials (ores, wood and stone) and consumables (food and water), ensuring that trading focuses on foundational resources while agents independently learn other mechanics such as crafting and enchantments. 

\paragraph{Agent Specializations} 
\textit{Craftax-Coop} features 3 agents, each assigned a unique specialization: Miner, Forager, or Warrior, with success in the environment requiring all three specializations to cooperate.

The Miner is tasked with mining for base materials (ores and stone), which they can then trade to other agents. They alone can craft pickaxes, needed to collect advanced materials (stone and ores). Similarly, only the Miner can craft and place torches, which are necessary for visibility and exploration in the dungeon layers. Finally, Miners also reserve the ability to place stones, needed when constructing shelter for agents to sleep.

The Forager is tasked with gathering food and water needed for agents' sustenance. Only Foragers can hunt passive mobs for food and interact with water sources (lakes and fountains), as well as plant and harvest crops. They also have an increased capacity to store food and water and must share these resources with other agents to avoid Miners and Warriors dying from thirst and hunger. 

The Warrior is tasked with combating enemies. They have double the base damage of other specializations, and they alone can construct advanced-level swords (stone swords and above), further increasing their damage output. They can also collect bows and craft arrows, used to combat enemies at range.

While the specializations define each agent's primary role, all agents maintain access to core mechanics, including movement, managing health and energy, enchantments and the ability to improve attributes (strength, dexterity and intelligence). This flexibility ensures that agents can adapt to various challenges, but their specialized skills necessitate continual collaboration to advance through levels and complete all achievements. For example, to craft a diamond sword, a Warrior must request diamond from the Miner who would search and gather it from the environment. The Warrior can then use the diamond sword to defend teammates against enemies of increased difficulty in later levels.

\paragraph{Health} 
As in Craftax, agents lose health if they fail to gather essential resources (food, water and energy) or if they take damage from hostile enemies, and die when their health points drop to zero. Additionally, agents can lose health through friendly-fire from other teammates. Agents can also revive others by approaching the dead agent and performing the \code{DO} action. The revived agents retain their inventory and their health is restored to one point. An episode terminates only when all agents' health reaches zero.

\subsection{RL Environment Interface}
Both \textit{Craftax-MA} and \textit{Craftax-Coop} conform to the JaxMARL interface \citep{rutherford2024jaxmarl}, facilitating easy integration with existing MARL algorithms and fast experimentation.

\paragraph{Observation Space} As in Craftax, both \textit{Craftax-MA} and \textit{Craftax-Coop} provide options for pixel-based and symbolic observations.
% Pixel-based observations consist of RGB images representing the environment from the agent's perspective, while symbolic observations encode the environment using compact discrete representations that are more computationally efficient to process.
Each agent only sees their local observation (consisting of their immediate surroundings and own inventory), meaning the environments can be modeled as a Dec-POMDP. To accommodate the heterogeneous agents in \textit{Craftax-Coop}, symbolic observations use one-hot encoding to represent different teammates in the agent's visual area, while the pixel observations render each teammate with an identifying shirt color. Each agent's own specialization is provided through one-hot encoding in the symbolic observations and an identifying icon in the pixel observations. Additional information is added in \textit{Craftax-Coop} to indicate each teammates' health, specialization, direction when off-screen and requested resource for trading. The pixel-based observation space is a $110 \times 130 \times 3$ image for \textit{Craftax-MA} (in line with Craftax), and a $110 \times 150 \times 3$ image for \textit{Craftax-Coop}. The symbolic observation space is of size 8465 for \textit{Craftax-MA}, and of size 8728 for \textit{Craftax-Coop}. A full description of the observation space is given in Appendix B.

\paragraph{Action Space}
Both environments maintain all 53 discrete actions from Craftax. To facilitate trading, \textit{Craftax-Coop} introduces a set of \code{REQUEST\_\{RESOURCE\}} actions which broadcast the request of the specified resource to all other agents for 10 timesteps. To provide the requested material, an agent performs a \code{GIVE\_\{AGENT\_I\}} action within the 10 timesteps after the first agent initiated the request. With regards to dead agents, their actions are replaced with no-ops, and in \textit{Craftax-Coop} they can be revived by others using the general \code{DO} action. A full description of the action space is given in Appendix C.

\paragraph{Reward and Achievements}
We follow a similar reward structure to Crafter and Craftax. Agents receive a reward the first time each agent completes an achievement each episode. All rewards are shared among the agents, regardless of which agent completes the achievement. Considering that player specializations limit the actions of certain agents, some achievements are limited to particular specializations. For example, only the Miner can craft a stone pickaxe and hence only they can collect the \code{MAKE\_STONE\_PICKAXE} achievement. This restriction of specialization capabilities also means that some achievements require collaboration among agents to achieve. For example, for a Warrior to collect the \code{MAKE\_STONE\_SWORD}, they must request stone from the Miner and use it to craft the sword. A full description of the achievements is given in Appendix D.

\subsection{Speed Evaluation}
\begin{figure}[t]
    \centering
    \includegraphics[width=0.5\textwidth]{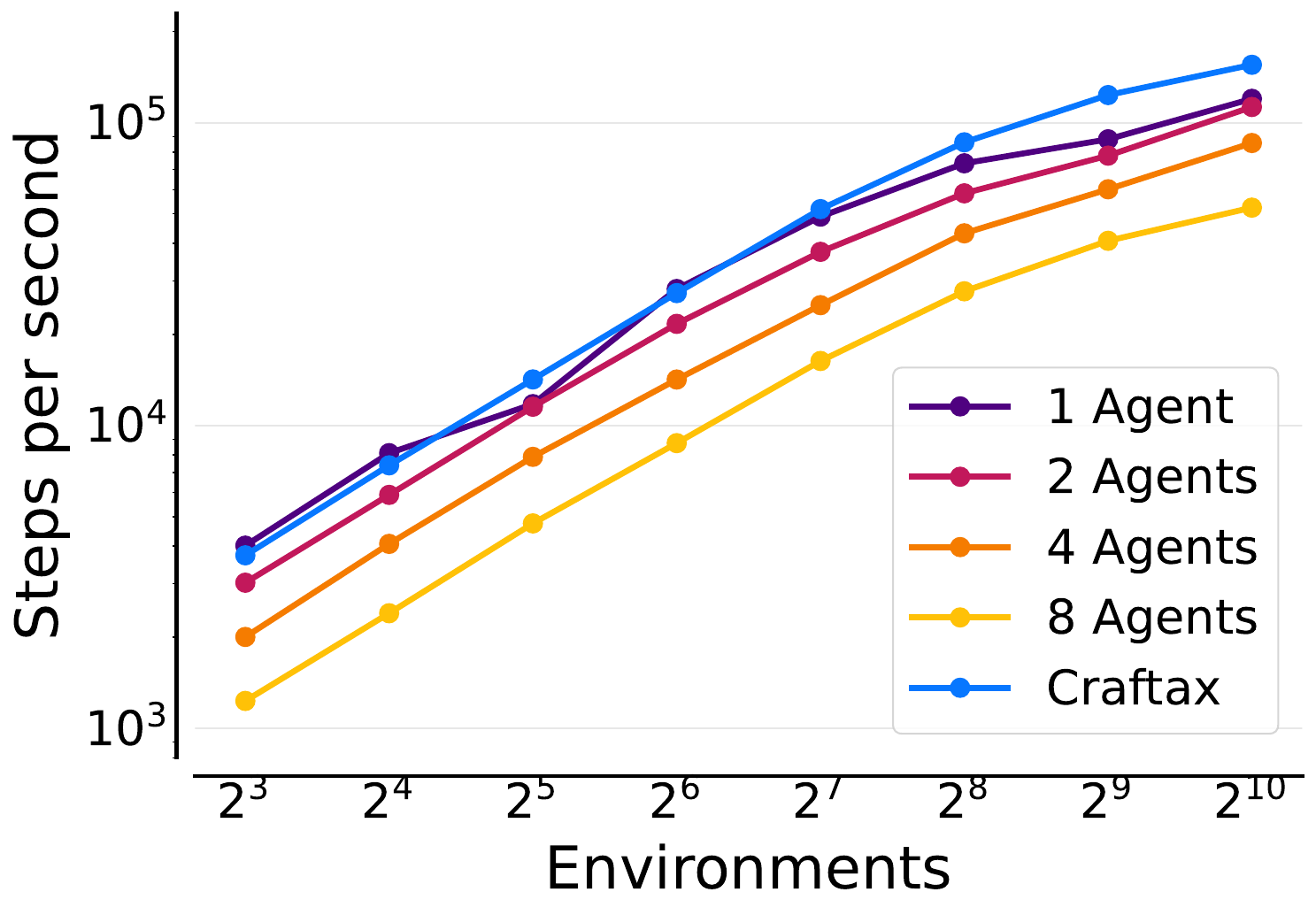}
    \caption{Analysis of \textit{Craftax-MA}'s ability to scale to thousands of parallel environments and different agent population counts. All measurements were recorded while training IPPO on a single L40S GPU. Results are compared to training PPO on Craftax. Scaling the number of parallel environments is nearly log-log linear with training throughput, while increasing the number of agents monotonically reduces the training throughput.}
    \label{fig:sps}
\end{figure}

Written in JAX, \textit{Craftax-MA} and \textit{Craftax-Coop} can be integrated with JaxMARL, enabling end-to-end hardware-accelerated benchmarking of MARL algorithms in our environments. This integration also leverages JAX's vectorization capabilities, enabling efficient scaling of agent populations and environments running in parallel (Figure \ref{fig:sps}). Using a single L40S GPU, a training run of IPPO with 4 agents in \textit{Craftax-MA} covers 250 million environment steps in 57 minutes. Similarly, a training run of IPPO with 3 agents in \textit{Craftax-Coop} covers 250 million environment steps in 52 minutes.

\section{Experiments and Results}
\subsection{Experimental Setup}
Our primary baseline is Multi-Agent Proximal Policy Optimization (MAPPO) \citep{yu2022mappo}, an extension of PPO \citep{schulman2017ppo} modified for the multi-agent setting by training a value network conditioned on global observations and a policy network conditioned on individual agent observations. In addition to MAPPO, we evaluate two other algorithms in the \textit{Craftax-Coop} environment: Independent PPO (IPPO), where each agent independently learns using PPO without centralized value estimation, and Parallelized Q-Network (PQN) \citep{gallici2025pqn}, an efficient Q-learning algorithm using parallel environments for scalable training.

To account for partial observability, we integrate memory into the networks using a Gated Recurrent Unit \citep{chung2014gru} for MAPPO and IPPO, and a Long-Short Term Memory Unit \citep{hochreiter1997lstm} for PQN. For multi-agent coordination with PQN, we use Value Decomposition Networks (VDN) \citep{sunehag2018vdn} to optimize the joint action-value function as the sum of individual agents' action-values.

For MAPPO and IPPO, we use hyperparameters identical to those used for PPO in \citet{matthews2024craftax}, with a decrease in environment workers for MAPPO to avoid memory limitations. For PQN, we use hyperparameters similar to those used in \citet{gallici2025pqn} for the Craftax environment. The full list of selected hyperparameters are listed in Appendix E. JaxMARL implementations of all algorithms were used to conduct these experiments.  Each algorithm is allocated a budget of 1 billion environment interactions, allowing sufficient opportunity for exploration, continual learning, long-term planning cooperation. For \textit{Craftax-MA}, we report the average rewards obtained per agents, and for \textit{Craftax-Coop} we report the total agent rewards. Episode rewards are reported as a percentage of the maximum achievable reward, which is 226 per agent for \textit{Craftax-MA} and 581 for all three agents for \textit{Craftax-Coop}.
\subsection{Craftax-MA}
\begin{figure}
    \centering
    \begin{subfigure}{0.45\textwidth}
        \includegraphics[width=\linewidth]{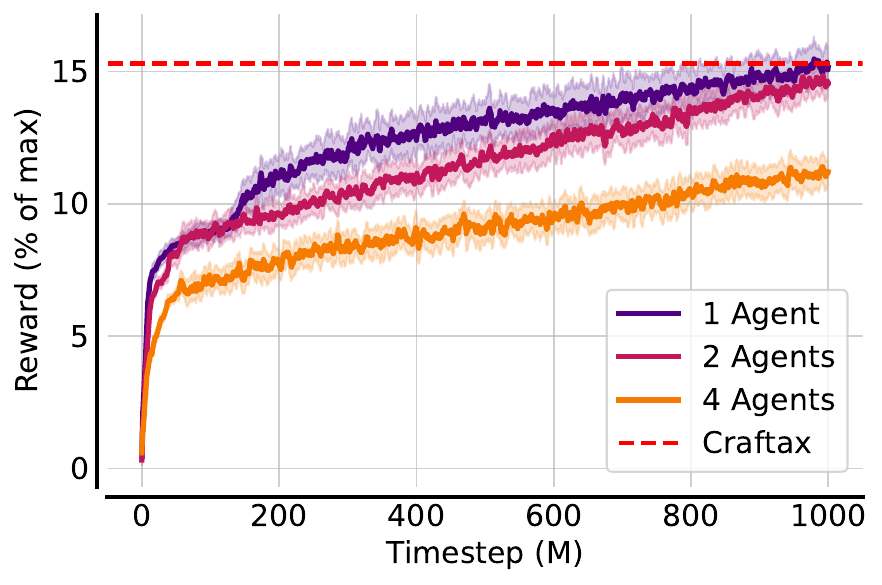}
        \caption{Shared Rewards}
        \label{fig:shared_rewards}
    \end{subfigure}
    \hfill
    \begin{subfigure}{0.45\textwidth}
        \includegraphics[width=\linewidth]{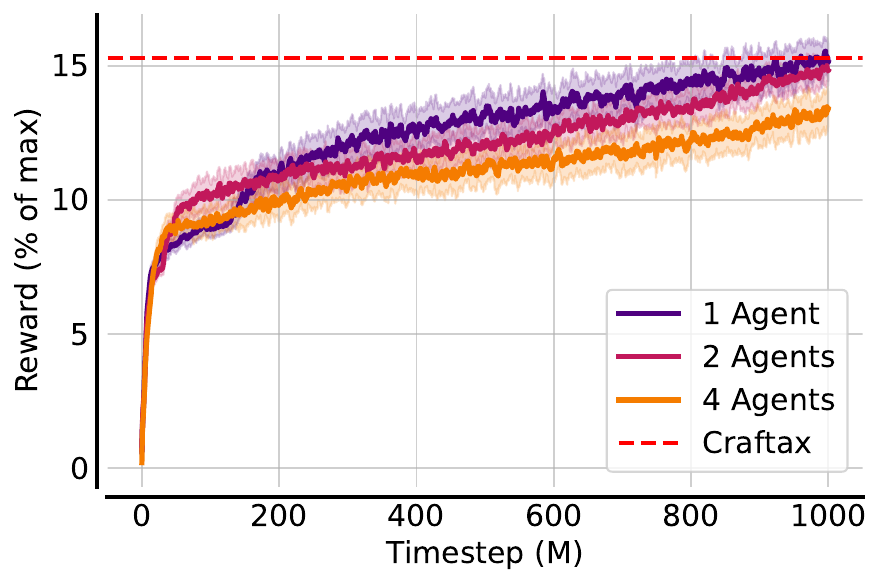}
        \caption{Individual Rewards}
        \label{fig:individual_rewards}
    \end{subfigure}
    % \begin{subfigure}{0.32\textwidth}
    %     \includegraphics[width=\linewidth]{assets/collect_resources_bar.pdf}
    %     \caption{Resource Collection Rate}
    %     \label{fig:resource_comparison}
    % \end{subfigure}
    \caption{Comparison of training performance of MAPPO in \textit{Craftax-MA} with \textbf{(a)} shared rewards and \textbf{(b)} individual rewards, for increasing number of agents.
    % \textbf{(c)} highlights the collection rate of resources after training MAPPO using 1 billion environment interactions in the individual reward setting.
    Results are also compared with the final reward of PPO-RNN on Craftax-1B \citep{matthews2024craftax}. Increasing the number of agents produces a decrease in the obtained returns, but a narrower difference in returns is observed under the individual reward setting.
    The experiments were repeated for 3 seeds, with the shaded area and error bars denoting 1 standard error.}
    \label{fig:reward_comparison}
\end{figure}

% \begin{wrapfigure}{r}{0.45\textwidth}
% \centering
% \includegraphics[width=\linewidth]{assets/collect_resources_bar.pdf}
% \caption{Collection rate of resources after training MAPPO using 1 billion environment interactions in the individual rewards setting. Experiments were repeated for 3 seeds, with the shaded area and error bars denoting 1 standard error.}
% \label{fig:resource_comparison}
% \end{wrapfigure}

In this section, we benchmark the performance of MAPPO with increasing populations of agents in \textit{Craftax-MA}. We scale the number of mobs (both enemies and passive mobs) with the number of agents to maintain a consistent challenge. For all settings we observe limited performance, with MAPPO for all agent population counts producing less than 15\% of the total reward. Under the default shared reward setting, we observe a decrease in the episodic returns as we increase the number of agents (Figure \ref{fig:shared_rewards}). This can be attributed to the noisy credit assignment present with shared rewards, as we observe a narrower difference in episodic returns under the individual reward setting (Figure \ref{fig:individual_rewards}). This trend can also be attributed to the increase in competition over available resources, as we observe a similar decrease in \code{COLLECT\_FOOD}, \code{COLLECT\_DRINK}, \code{COLLECT\_STONE} and other resource related achievements obtained in the individual reward settings (Figure \ref{fig:resource_comparison}).

\subsection{Craftax-Coop}
\begin{figure}[t]
    \begin{minipage}{0.45\textwidth}
        \centering
        \includegraphics[width=\linewidth]{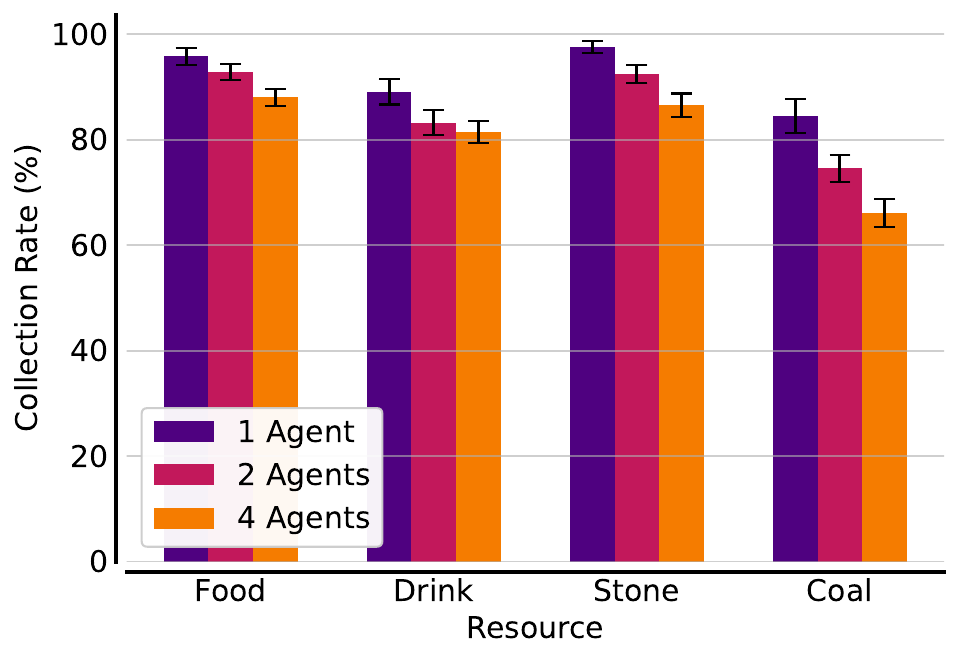}
        \caption{Collection rate of resources after training MAPPO using 1 billion environment interactions in the individual rewards setting. As the number of agents is increased, the collection rate of resources consistently decrease. Experiments were repeated for 3 seeds, with the shaded area and error bars denoting 1 standard error.}
        \label{fig:resource_comparison}
    \end{minipage}
    \hfill
    \begin{minipage}{0.45\textwidth}
        \centering
        \includegraphics[width=\linewidth]{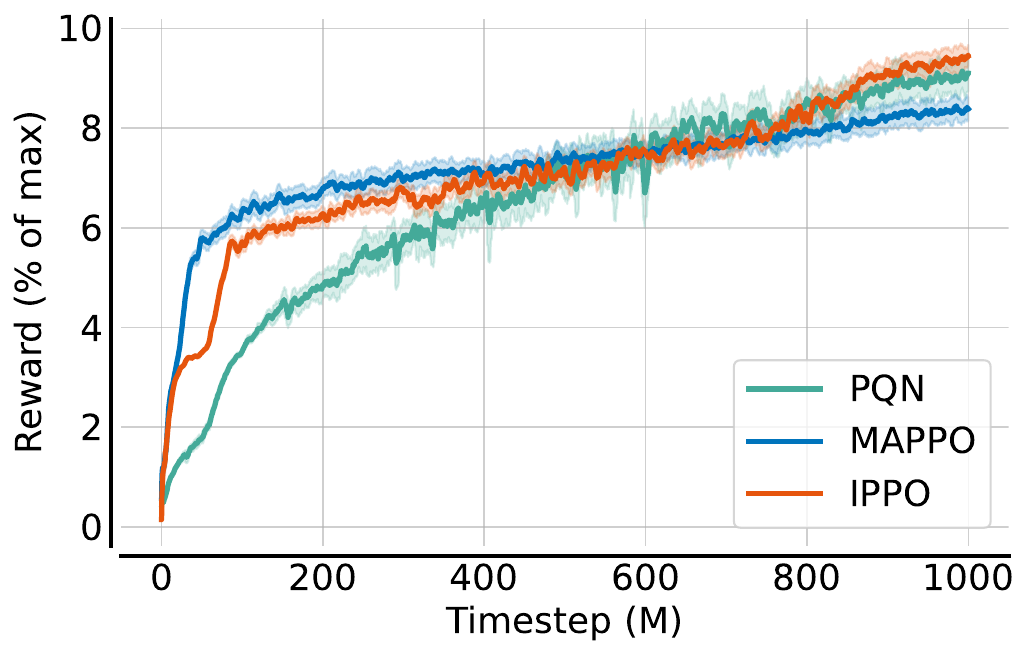}
        \caption{Performance comparison of MAPPO, IPPO PQN on the \textit{Craftax-Coop} environment with 3 agents. MAPPO produces the lowest final episodic returns compared to other algorithms. Each algorithm is run for 1 billion timesteps with 3 seeds. The shaded area denotes 1 standard error.}
        \label{fig:base_coop_plot}
    \end{minipage}\hfill
\end{figure}
\begin{figure}
    \centering
    \begin{subfigure}{0.45\textwidth}
        \includegraphics[width=\linewidth]{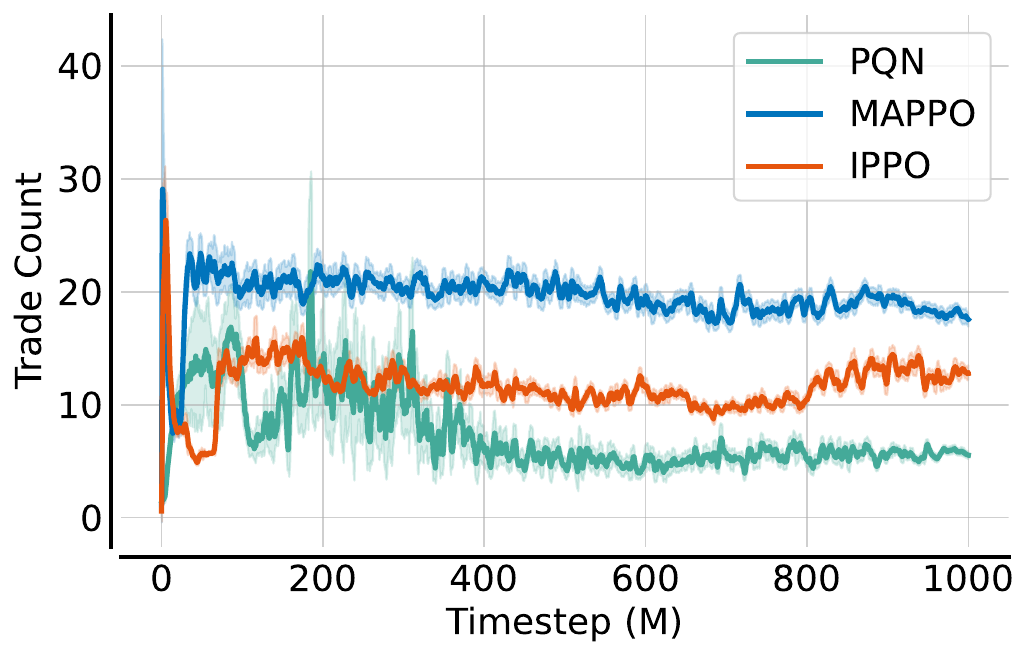}
        \caption{Trade Count}
        \label{fig:trade_count}
    \end{subfigure}%
    \hfill
    \begin{subfigure}{0.45\textwidth}
        \includegraphics[width=\linewidth]{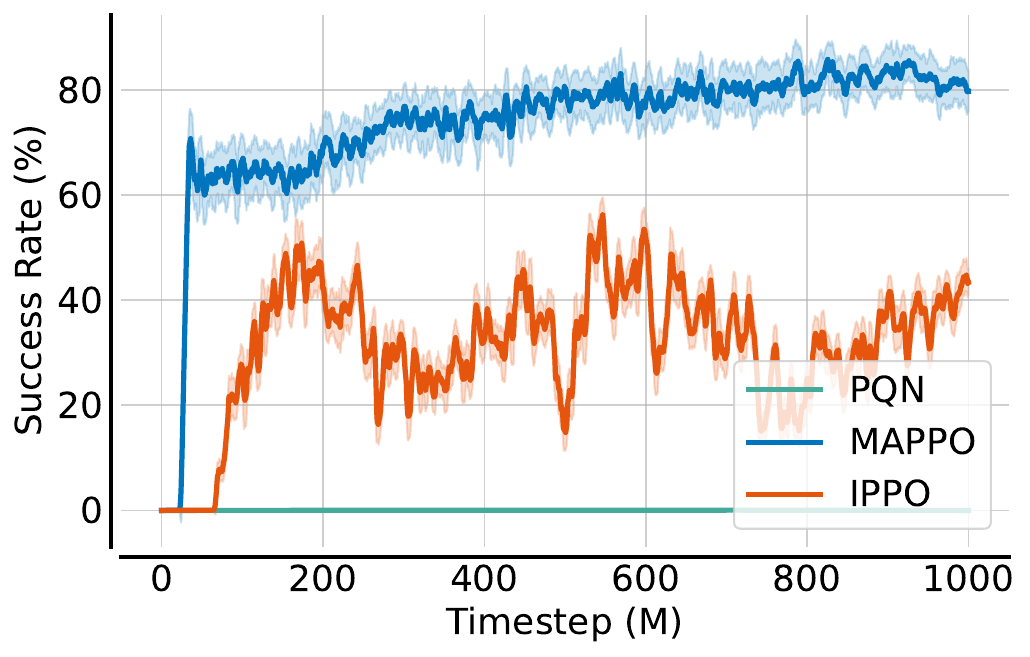}
        \caption{Stone Sword Collection Rate}
        \label{fig:stone_sword}
    \end{subfigure}
    \caption{Performance comparison of MAPPO, IPPO PQN in \textit{Craftax-Coop} with 3 agents in (a) average trades per episode and (b) average stone sword crafting rate. Increased trading is observed with MAPPO compared to IPPO and PQN, leading to a higher stone sword crafting rate which depends on the stone trading between the Miner and the Warrior. Algorithms trained for 1 billion timesteps with 3 seeds. Shaded area denotes 1 standard error.}
    \label{fig:full_base_results}
\end{figure}
In this section, we benchmark the performance of MAPPO, IPPO and PQN against the \textit{Craftax-Coop} environment with 3 agents. The returns for the evaluated algorithms are summarized in Figure \ref{fig:base_coop_plot}, with the fine-grained achievement results shown in Appendix F. We observe comparable final performance across all three algorithms, with MAPPO recording the lowest final episodic return despite being the fastest to learn at first.

\paragraph{Cooperation}
The player specializations featured in \textit{Craftax-Coop} require cooperation among agents for success in the environment, and among these cooperation challenges is resource sharing. We observe in \ref{fig:trade_count} the most trading among agents occurs with MAPPO, followed by IPPO and then PQN. We observe in Figure \ref{fig:stone_sword}, that the Warrior is able to craft stone swords, highlighting that agents are learning to perform meaningful resource sharing, such as the Miner trading gathered stone to the Warrior. This behavior is mainly observed with MAPPO, but less so with IPPO and not at all with PQN, highlighting a gap between these algorithms in the ability to cooperate.

\paragraph{Credit Assignment}
Several challenges in \textit{Craftax-Coop} require long-term reasoning to overcome. Among these is collecting and distributing food and water between agents to maintain their health. The change of health associated with gathering or not gathering food and water is only observed many timesteps later, making this a challenging temporal credit assignment problem. Upon qualitatively analyzing episodes of fully trained agents, we observe that majority of agents die from thirst and hunger. We ablate adding an incentive for maintaining food and water in each agent's inventory. A reward of -0.1 is given for every unit of food or water lost and vice versa. As the maximum amount of food and water each agent can obtain is capped, this reward signal cannot be exploited by continually gathering these resources. However, it does provide an immediate reward upon gathering or losing food and water, simplifying the associated temporal credit assignment problem. 

\begin{figure}
    \centering
    \begin{subfigure}[b]{0.45\textwidth}
        \centering
        \includegraphics[width=\textwidth]{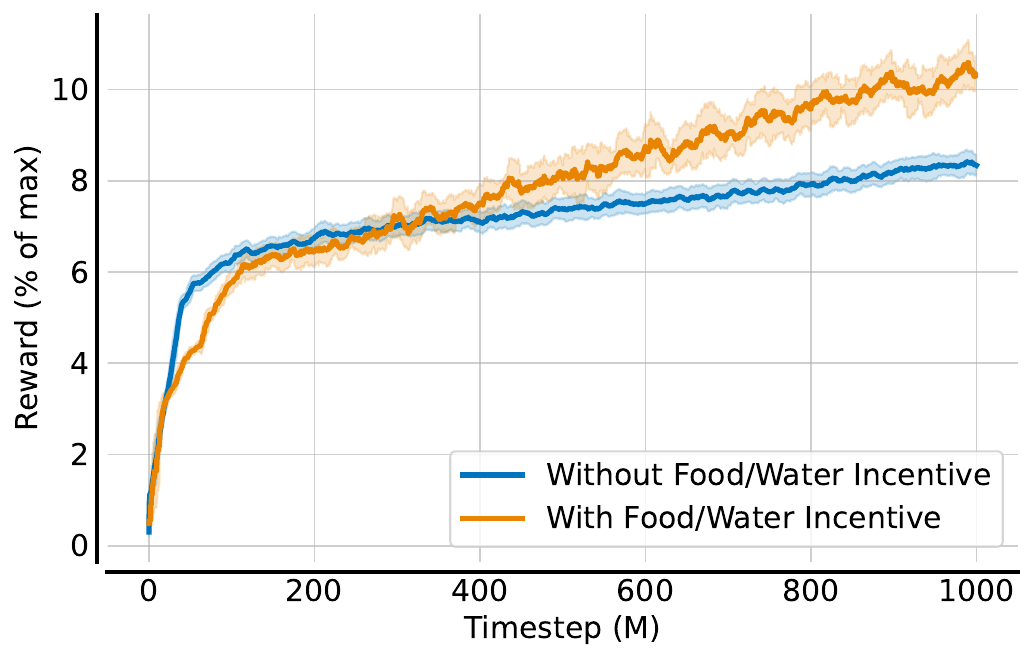}
        \caption{Performance with food/water incentive}
        \label{fig:nofoodwater}
    \end{subfigure}
    \hfill
    \begin{subfigure}[b]{0.45\textwidth}
        \centering
        \includegraphics[width=\textwidth]{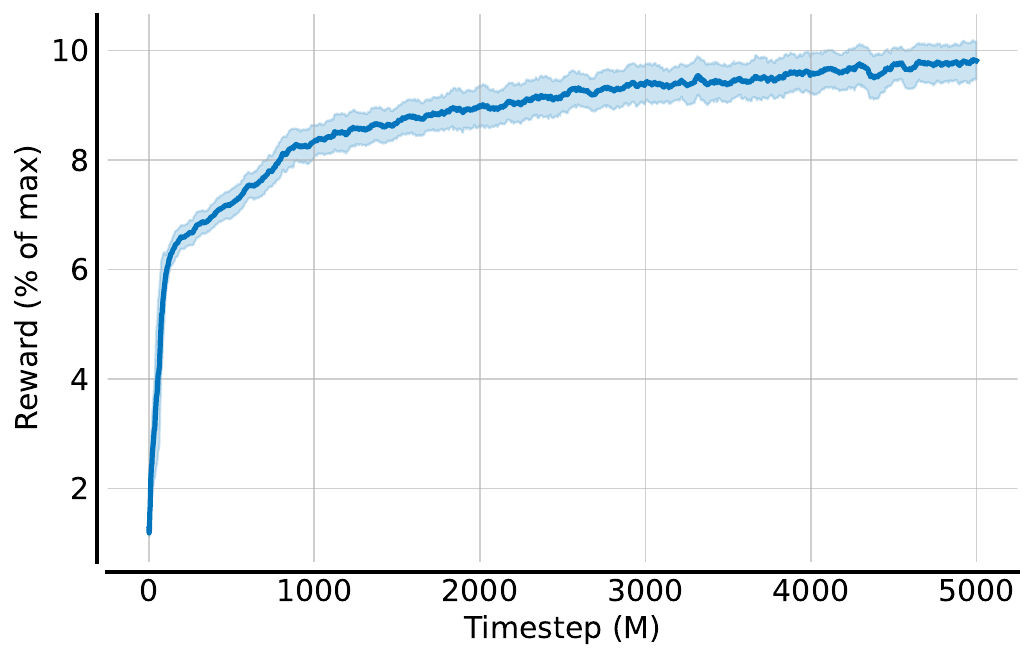}
        \caption{Performance over 5 billion steps}
        \label{fig:exploration_5b}
    \end{subfigure}
    \caption{
        Training performance of MAPPO in \textit{Craftax-Coop} with 3 agents: (a) with an additional incentive to maintain food and water, leading to noticeably increased performance compared to the base environment; and (b) over 5 billion steps, where learning stagnates after 1 billion timesteps and increases by only 2\% in the following 4 billion steps. Both experiments were repeated for 3 seeds and the shaded area denotes 1 standard error.
    }
    \label{fig:combined_figure}
\end{figure}

As seen in Figure \ref{fig:nofoodwater}, we observe that the episodic returns in this setting increase by a noticeable margin. This ablation demonstrates the large value provided by overcoming the credit assignment problem associated with collecting food and water, which current MARL algorithms struggle to do. Full experiment results are presented in Appendix G.

\paragraph{Exploration}
\textit{Craftax-Coop} provides a significant exploration challenge due to its multi-level world and extensive reward hierarchy. We demonstrate that existing MARL algorithms struggle with exploration in this environment by extending training of MAPPO to 5 billion environment steps. As shown in Figure \ref{fig:exploration_5b}, the algorithm performance stagnates at around 1 billion steps, and only increases by 2\% of total rewards in the 4 billion steps that follow. At the end of the 5 billion steps, less than 2\% of episodes have agents reach the 3rd level (Gnomish Mines) of all 9 levels in the environment. These results highlight significant exploration limitations in current MARL algorithms, and we believe \textit{Craftax-Coop} can serve as a benchmark for future progress here. Full experiment results are presented in Appendix H.

\section{Related Work}

Our work fits into a diverse literature of existing environments including those that are multi-agent, JAX-based and that focus on open-endedness.

\subsection{Cooperative Multi-Agent Benchmarks}

Multi-agent benchmarks have a long history, with early work including Keepaway Soccer \citep{stone2006keepaway} and Half Field Offense \citep{kalyanakrishnan2007half}, both implemented in the RoboCup simulator \citep{kitano1997robocup}. QMIX \citep{rashid2020monotonic} was released along with the highly influential Starcraft Multi-Agent Benchmark \citep{samvelyan2019starcraft}, where many agents have to work together co-operatively to defeat an enemy team. Problems involving stochasticity and partial observability were later dealt with in SMACv2 \citep{ellis2023smacv2}.  Other popular benchmarks include Hanabi \citep{bard2020hanabi},  Level Based Foraging \citep{christianos2020shared} and Google Football \citep{kurach2020google}. While these environments have facilitated the development of increasingly complex MARL algorithms, they are generally quite slow to run, limiting their applicability for researchers. Furthermore, these environments all operate on relatively short time horizons, with episodes commonly lasting for less than a thousand timesteps. In contrast, a successful run of Multi-Agent Craftax will typically take tens of thousands of timestep, with early decisions having a large effect later in the episode, allowing for the study of long-term dependencies.

\subsection{Hardware Accelerated Environments}

There has been a recent explosion of hardware accelerated environments for both single and multi-agent RL. Prominent single agent examples include Gymnax \citep{gymnax2022github}, which reimplements many classic RL environments such as Cartpole \citep{barto1983neuronlike} and MinAtar \citep{young2019minatar}; Brax \citep{freeman2021brax}, which simulates physical locomotion problems in the style of MuJoCo \citep{todorov2012mujoco}; XLand-Minigrid \citep{nikulin2024xland, nikulin2024xland100b}, which implements a simplified version of the XLand simulator \citep{team2021open} and Jumanji \citep{bonnet2023jumanji}, which contains many combinatorial problems.

Examples of multi-agent environments include the JaxMARL suite of environments \citep{rutherford2024jaxmarl}; Pgx \citep{koyamada2023pgx}, which contains competitive multi-agent games like Chess and Go; GPUDrive \citep{kazemkhani2024gpudrive}, which simulates the training of autonomous vehicles and VMAS \citep{bettini2022vmas}, which contains a set of multi-robot tasks.

Compared to traditional MARL benchmarks, these hardware-accelerated environments are even more skewed towards short-horizon tasks.

\subsection{Environments for Open-Endedness}

Our work is also related to the burgeoning field of open-endedness \citep{stanley2017openendedness}, in which learning occurs continually with ever increasing complexity. Existing environments in this category include MALMO \citep{johnson2016malmo}, the NetHack Learning Environment \citep{kuttler2020nethack}, Crafter \citep{hafner2021benchmarking}, XLand-Minigrid \citep{nikulin2024xland} and Kinetix \citep{matthews2024kinetix}.

Perhaps the work most similar to ours is Neural MMO \citep{suarez2019neural}, a massively multi-agent and open-ended environment in which agents compete for resources in a general-sum environment. Recent integrations with PufferLib \citep{suarez2024pufferlib} have also given Neural MMO an impressive speed of experimentation.  Despite their clear similarity, we see many differences between our own work and Neural MMO. Firstly, whereas most of the complexity in Neural MMO arises from interactions between the agents, the base Craftax game mechanics form the static and challenging complexity for our own work. The focus of Neural MMO is on emergent multi-agent phenomena that occurs from simulating large populations of agents, whereas our work presents a challenging task that a handful of agents need to work together to solve.

\citet{ye2024efficient} also implement a multi-agent version of the simpler Craftax-Classic benchmark (analogous to the original Crafter environment), while we focus on adapting the significantly more challenging main Craftax benchmark.

\section{Conclusion}
We present a multi-agent extension to the popular Craftax benchmark, through the \textit{Craftax-MA} environment. We further extend this environment, adding agent heterogeneity, trading and other mechanics to provide a compelling cooperation challenge for MARL through \textit{Craftax-Coop}. The two environments provide a scalable and efficient platform for studying MARL, focusing on cooperation, exploration, long-term planning and credit assignment. Our experiments demonstrate that popular MARL adaptations of algorithms struggle with these challenges, highlighting the potential of our benchmark to drive future development of more capable and cooperative agents.

\paragraph{Limitations and Future Work}
While \textit{Craftax-MA} and \textit{Craftax-Coop} provide a robust platform for MARL research, our experiments are limited to relatively small populations of agents. Future work should explore the scalability of agent populations in these environments beyond just four agents, testing the ability of algorithms to manage large-scale cooperative interactions. Additionally, we plan to integrate text rendering capabilities to facilitate the evaluation of large language model (LLM) agents in the environment, allowing for a more direct assessment of their ability to plan, cooperate explore in complex, dynamic, multi-agent settings.

\bibliographystyle{unsrtnat}
{%\small
\bibliography{references}}

\end{document}